\documentclass{bmvc2k}

\usepackage{times}
\usepackage{epsfig}
\usepackage{graphicx}
\usepackage{amsmath}
\usepackage{amssymb}
\usepackage{multirow}
\usepackage{algorithm}
\usepackage{algorithmicx}
\usepackage{algpseudocode}
\usepackage{multicol}
\usepackage{hyperref}

\floatname{algorithm}{Algorithm}

\newcommand{\figref}[1]{Fig.~\ref{#1}}
\newcommand{\tabref}[1]{Tab.~\ref{#1}}

\newcommand{\secref}[1]{Sec.~\ref{#1}}
\newcommand{\algoref}[1]{Alg.~\ref{#1}}

\graphicspath{{./Figures/}}

\title{An Evaluation of Feature Matchers for \\ Fundamental Matrix Estimation}

\addauthor{Jia-Wang Bian$^{\text{1,}}$}{jiawang.bian@adelaide.edu.au}{2}
\addauthor{Yu-Huan Wu}{wuyuhuan@mail.nankai.edu.cn}{3}
\addauthor{Ji Zhao}{ji.zhao@tusimple.ai}{4}
\addauthor{Yun Liu}{nk12csly@mail.nankai.edu.cn}{3}
\addauthor{Le Zhang}{zhangleuestc@gmail.com}{5}
\addauthor{Ming-Ming Cheng}{cmm@nankai.edu.cn}{3}
\addauthor{Ian Reid$^{\text{1,}}$}{ian.reid@adelaide.edu.au}{2}

\addinstitution{
School of Computer Science, \\
The University of Adelaide, \\
Adelaide, Australia
}
\addinstitution{
Australian Centre for Robotic \\Vision,
Australia
}
\addinstitution{
Nankai University, 
China
}
\addinstitution{
 TuSimple, China \\
}
\addinstitution{
Agency for Science, Technology \\
and Research, Singapore
}

\runninghead{Bian et al.}{An Evaluation of Feature Matchers}

\def\eg{\emph{e.g.}}
\def\ie{\emph{i.e.}}
\def\etal{\emph{et al.}}

\begin{document}
\maketitle

\begin{abstract}
Matching two images while estimating their relative geometry is a key step in many computer vision applications.
For decades, a well-established pipeline, consisting of SIFT, RANSAC, and 8-point algorithm, has been used for this task.
Recently, many new approaches were proposed and shown to outperform previous alternatives on standard benchmarks, including the learned features, correspondence pruning algorithms, and robust estimators.
However, whether it is beneficial to incorporate them into the classic pipeline is less-investigated.
To this end, we are interested in \textbf{i)} evaluating the performance of these recent algorithms in the context of image matching and epipolar geometry estimation,
and \textbf{ii)} leveraging them to design more practical registration systems.
The experiments are conducted in four large-scale datasets using strictly defined evaluation metrics, 
and the promising results provide insight into which algorithms suit which scenarios.
According to this, we propose three high-quality matching systems and a Coarse-to-Fine RANSAC estimator.
They show remarkable performances and have potentials to a large part of computer vision tasks.
To facilitate future research, the full evaluation pipeline and the proposed methods are made publicly available.
\end{abstract}

%-------------------------------------------------------------------------
\section{Introduction}

Matching two images while recovering their geometric relation, \eg, \emph{epipolar geometry}~\cite{hartley2003multiple}, is one of the most basic tasks in computer vision and a crucial step in many applications such as Structure-from-Motion 
(SfM)~\cite{agarwal2011building,heinly2015reconstructing,radenovic2016dusk,schonberger2015single,schonberger2016structure} and Visual SLAM~\cite{mur2015orb,davison2007monoslam,forster2014svo}.
In these applications, the overall performance heavily depends on the quality of the initial two-view registration.
Consequently, a thorough performance evaluation for this module is of vital importance to the computer vision community.
However, to the best of our knowledge, no previous work has done it.
To this end, we are dedicated to an extensive experimental evaluation of existing algorithms to establish a uniform evaluation protocol in this paper.

\begin{figure}[t]
\centering
\includegraphics[width=0.8\linewidth]{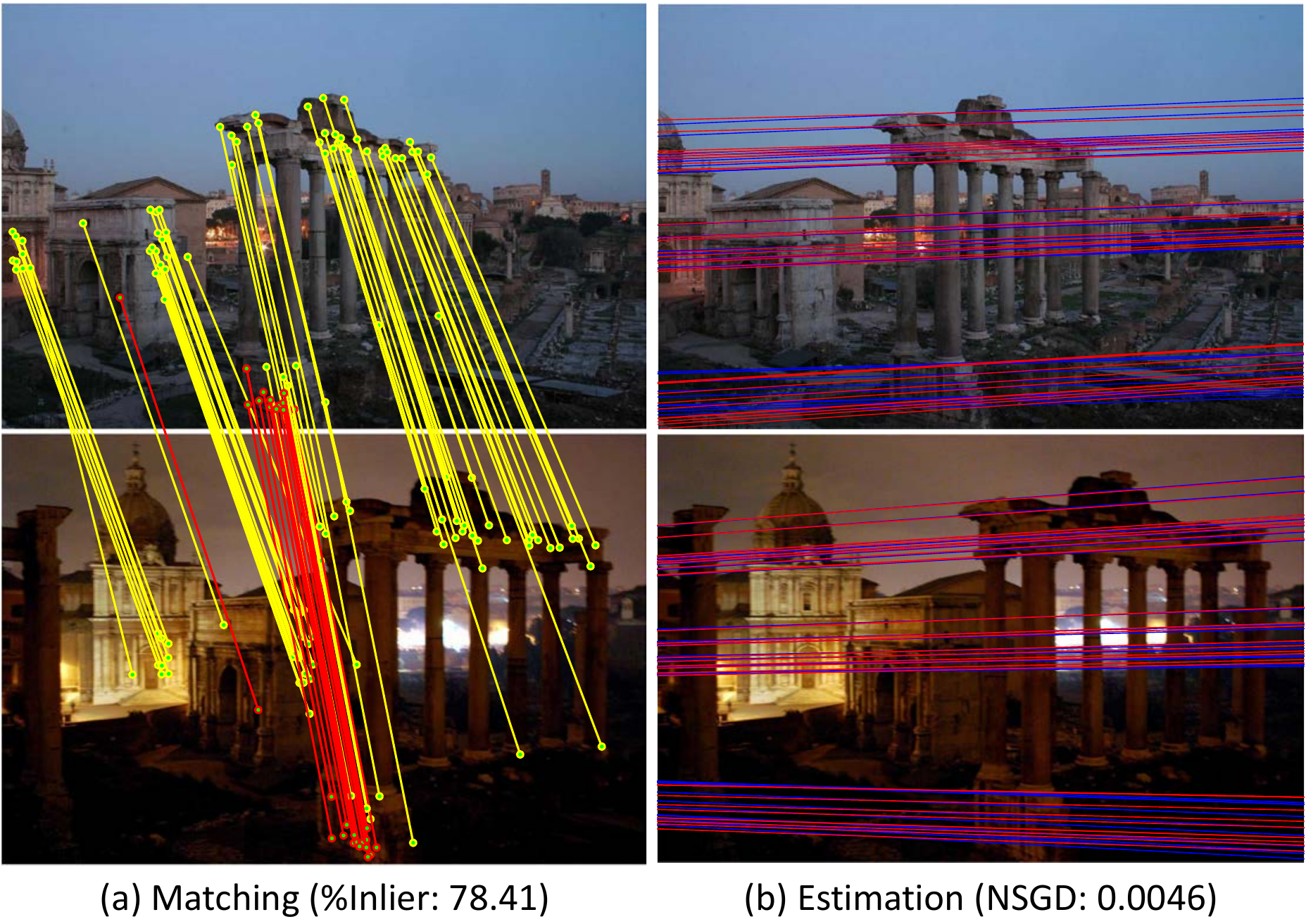}
\caption{
Outputs of the two-view matching and geometry estimation pipeline.
(a) shows matching results with yellow/red lines standing for inliers/outliers.
(b) shows epipolar geometry estimation results with red/blue lines standing for the ground-truth/estimated epipolar lines.
We use the proposed \emph{normalized symmetric geometric distance} (NSGD) to measure the estimation accuracy.
The smaller, the better.
}
\label{fig-pipeline}
\vspace{-3mm}
\end{figure}

For decades, a classic pipeline has been used for this task, which relies on the SIFT~\cite{lowe2004distinctive} features to establish initial correspondences across images, then prunes bad correspondences by Lowe's \emph{ratio test}~\cite{lowe2004distinctive}, and finally estimates the geometry using RANSAC~\cite{fischler1981random} based estimators.
We are here interested in recovering the \emph{fundamental matrix} (FM), which suits more general scenes than other geometric models, \eg, the \emph{homography} and \emph{essential matrix}.
\figref{fig-pipeline} shows an example output of this pipeline.
Here, we mainly focus on the geometry estimation quality.

Recently, many new approaches were proposed which showed potentials to this task, including the learned features~\cite{mishchuk2017working,luo2018geodesc,via2018repeatability}, robust estimators~\cite{raguram2013usac,barath2018graph}, and, especially, correspondence pruning algorithms~\cite{bian2017gms,Ma2018,yi2018learning} which revived comparatively little attention over before.
However, while these algorithms outperform earlier ones on standard benchmarks, incorporating them into the classic pipeline may not necessarily translate into a performance increase.
For example, Balntas \etal~\cite{balntas2016learning} showed that descriptors which perform better than others on the standard benchmark~\cite{brown2011discriminative} do not show a better image matching quality.
The inconsistency was also shown and discussed in~\cite{bian2017gms,yi2018learning,schonberger2017comparative}.

In this paper, we conduct a comprehensive evaluation of recently proposed algorithms by incorporating them into the well-established image matching and epipolar geometry estimation pipeline to investigate whether they can increase the overall performance.
In detail, this paper makes the following contributions:

\begin{itemize}
\setlength\itemsep{0em}
\item \textbf{i)} We present an evaluation protocol for local features, robust estimators, and especially correspondence pruning algorithms such as~\cite{bian2017gms,yi2018learning,Ma2018} which have not been carefully investigated.

\item \textbf{ii)} We evaluate algorithms on four large-scale datasets using strictly defined metrics.
The results provide insights into which datasets are particularly challenging and which algorithms suit which scenarios.

\item \textbf{iii)} Based on the results, we propose three high-quality and  efficient matching systems, which perform on par with the powerful CODE~\cite{lin2017code} system but are several orders of magnitude faster.

\item \textbf{iv)} Interestingly, we observe that the recent GC-RANSAC~\cite{barath2018graph} (also USAC~\cite{raguram2013usac}) does not show consistently high performance on geometry estimation but permits effective outlier pruning.
We hence propose to first use it for outlier removal, and then apply LMedS based estimator~\cite{rousseeuw2005robust} for model fitting.
The resulting approach, termed Coarse-to-Fine RANSAC, shows significant superiority over other alternatives.

\end{itemize}

\section{Related work}~\label{sec-relatedwork}

Rich research focuses on evaluating local features and robust estimators, while correspondence pruning algorithms have not been well evaluated.
The proposed benchmark mitigates this gap.
% In the following, we discuss representative evaluation protocols and ours.

% \vspace{2mm}
% \noindent
% \textbf{Evaluating Local Features.}
\paragraph{Evaluating Local Features.}
Mikolajczyk \etal~\cite{mikolajczyk2005comparison} evaluated the affine region detectors on small-scale datasets, which cover various photometric and geometric image transformations.
Later, Mikolajczyk and Schmid~\cite{mikolajczyk2005performance} extended the evaluation to local descriptors.
Build upon this, Heinly \etal \cite{heinly2012comparative} proposed several additional metrics and datasets to evaluate binary descriptors.
Besides, Brown \etal~\cite{brown2011discriminative} presented a patch pair classification benchmark for the learned descriptors, which measures the ability of a descriptor to discriminate positive from negative patch pairs.
Recently, Balntas \etal \cite{hpatches_2017_cvpr} evaluated the hand-crafted and learned descriptors in terms of the verifying and retrieving \emph{homography} patches.
Sch\"{o}nberger \etal \cite{schonberger2017comparative} comparatively evaluated these two types of descriptors in the context of image-based reconstruction.

\paragraph{Evaluating Robust Estimators.}
Choi \etal \cite{choi1997performance} conducted an evaluation of RANSAC~\cite{fischler1981random} family in terms of the line fitting and homography estimation~\cite{hartley2003multiple}, where the accuracy, runtime, and robustness of methods are analyzed.
Lacey \etal \cite{lacey2000evaluation} performed an evaluation of RANSAC algorithms for stereo camera calibration.
Raguram \etal \cite{raguram2008comparative} categorized RANSAC algorithms and provide a comparative analysis on them, where the trade-off between efficiency and accuracy is considered.
These protocols evaluate robust model fitting techniques in both synthetic and real data.
Torr \etal~\cite{torr1997development,torr1997performance} provided performance characterization of \emph{fundamental matrix} (FM) estimation algorithms.
Zhang~\cite{Zhang1998} reviewed FM estimation techniques and proposed a well-founded measure to compute the distance of two fundamental matrices, which is shown to better than using the Frobenius norm.
Armangu\`{e} \etal~\cite{armangue2003overall} provided an overview on different FM estimation approaches.
Fathy \etal~\cite{fathy2011fundamental} studied the error criteria in FM estimation phase.

\paragraph{Proposed Benchmark.}
Our benchmark is mainly motivated by \cite{schonberger2017comparative} which evaluates descriptors in higher-level tasks.
The difference is that we evaluate three types of algorithms in the context of two-view image matching and geometry estimation for the overall performance, while \cite{schonberger2017comparative} evaluates descriptors in multiple tasks for the generalized descriptor.
Besides, we draw from~\cite{Zhang1998,ranftl2018deep,torr1997development,schonberger2017comparative,barath2018graph} to design the evaluation metric and construct the benchmark dataset.
Moreover, the presented evaluation could also be interpreted as an ablation study for image matching and geometry estimation pipeline.
It can help researchers design more practical correspondence systems.

\begin{algorithm}[t]
\caption{Compute SGD}
\small
\begin{multicols}{2}
\begin{algorithmic}[1]
\Require $I_1, I_2, F_1, F_2, N$
\Ensure $sgd$
\Function{ComputeSGD}{$I_1, I_2, F_1, F_2, N$}
	\State $sgd \gets  0$
    \State $count \gets  0$
    \While{$count < N$}
    	\State randomly choose a point $m$ in $I_1$
    	\State draw $l_1 = F_1m$ in $I_2$        
    	\If{$l_1$ does not intersect with $I_2$}
        	\State continue
        \EndIf
        \State randomly choose a point $m'$ on $L_1$
        \State draw $l_2 = F_2m$ in $I_2$
        \State $d_1'$ = distance($m', l_2$)
        \State draw $l_3 = F^{T}_2 m'$ in $I_1$
		\State $d_1$ = distance($m, l_3$)
        \State $sgd \gets sgd + d_1' + d_1$
        \State $count \gets count + 1$
     \EndWhile
     \State swap $(I_1, I_2)$, swap $(F_1, F_2)$
     \State repeat step $3-17$
     \State $sgd \gets sgd / (4 * N)$
     \State \Return{$sgd$}
\EndFunction
\end{algorithmic}
\end{multicols}
\label{algo-sgd}
\end{algorithm}

% \begin{figure*}[t]
% \centering
% \includegraphics[width=1.0\linewidth]{examples.pdf}
% \caption{Sample results with Normalized SGD. 
% Red/blue lines stand for the ground-truth/estimated epipolar lines.
% We use $0.05$ (the smaller, the better) as the threshold to classify estimates as accurate or not.
% }
% \label{fig-nsgd}
% \vspace{-3mm}
% \end{figure*}

\section{Evaluation metrics}~\label{sec-metrics}

% In the following, we present evaluation metrics on fundamental matrix (FM) estimation and image matching.

\subsection{Metrics on FM estimation}

Fundamental matrices cannot be compared directly due to their structures.
For measuring the accuracy of estimation, 
we follow Zhang's method~\cite{Zhang1998}, referred as \emph{symmetric geometry distance} (SGD) in this paper.
It generates virtual correspondences using the ground-truth FM and computes the \emph{epipolar distance} to the estimated one,
and then reverts their roles to compute the distance again to ensure symmetry.
The averaged distance is used for accuracy measurement.
\algoref{algo-sgd} presents an overview for the computation of the SGD error,
where ($I_1$, $I_2$) is an image pair, $F_1$ and $F_2$ are two FMs, and $N$ is the number of maximum iterations.

% we refer for two recent works:
% Barath \etal~\cite{barath2018graph} compute the \emph{Sampson} distance of the FM to manually labeled correspondences, and Ranftl \etal~\cite{ranftl2018deep} generate virtual correspondences using the ground-truth FM and compute the \emph{Epipolar} distance to the estimated one.
% Here, the former is more accurate than the latter but limits in large-scale datasets and not physically intuitive, so we prefer the second solution.
% Indeed, the algorithm by Zhang~\cite{Zhang1998}, which can compute the distance of two FMs, formally implements this idea, and enforces an additional cross-check to ensure symmetrical ness.
% We follow this method, named \emph{symmetric geometry distance} (SGD) in this paper, to analyze the estimation results.
% However, the computed error, \emph{in pixels}, causes comparability issues between images of different resolutions.
% Therefore, we propose the normalized measure for evaluation as follows.

\paragraph{Normalized SGD.} 
The computed SGD error (\emph{in pixels}) causes comparability issues between images with different resolutions.
In order to address this issue, we propose to normalize the distance into the range of $[0,1]$ by dividing the distance by the length of the image diagonal.
Formally, the distance is regularized by multiplying a factor $f = 1 / \sqrt{h^2 + w^2}$, where $h$ and $w$ stand for the height and width of the image, respectively.
This makes the error comparable across different resolution images.

\paragraph{\%Recall.}
Given the FM estimates, we classify them as accurate or not by thresholding the Normalized SGD error,
and use the \%Recall, the ratio of accurate estimates to all estimates, for evaluation.
In our experiments, $0.05$ is used as the threshold.
As the recall increasing with thresholds in an accumulative way, the performance is not sensitive to threshold selection.
However, we also suggest readers showing recall curves with varying thresholds.

% ~\figref{fig-nsgd} shows sample results, where the measure is visually meaningful (the smaller, the better), and the threshold can separate accurate and inaccurate estimates.
% Besides, as the recall increase accumulatively with the increasing accuracy threshold slightly shifting the threshold would not change the performance comparison of different algorithms. 
% Here, we do not emphasize the accuracy of estimates, because 
% \emph{i)} it does not matter how inaccurate false estimates are,
% and \emph{{ii)}} it is also not crucial how accurate correct estimates are, as which can be optimized significantly by the back-end of high-level applications, \eg, Bundle Adjustment~\cite{triggs1999bundle}.
% In contrast, the robustness of initial two-view registration system is vital in the high-level applications, \eg, SfM systems would fail to reconstruct the scene if insufficient pairs are registered.
% This coincides with the discussion in~\cite{schonberger2017comparative} that matching more image pairs is more important than providing more correspondences on matched pairs by SIFT~\cite{lowe2004distinctive} in image-based reconstruction~\cite{schonberger2016structure}.

\subsection{Metrics on Image Matching}

% The standard metrics by Mikolajczyk \etal~\cite{mikolajczyk2005performance}, Heinly \etal~\cite{heinly2012comparative}, and Brown \etal~\cite{brown2011discriminative} offer detailed analyses for feature extraction and matching.
% However, incorporating algorithms which achieve better performances on standard datasets into high-level applications may not increase overall system performance~\cite{balntas2016learning,schonberger2017comparative,bian2017gms,yi2018learning}.
% The recent research~\cite{schonberger2017comparative} shows that the precision (inlier rate) of correspondences is more practically important for geometry estimation and subsequent algorithms.
% We thus examine matching performance by measuring the inlier rate of outputted matches, and besides that, the match numbers are also analyzed.

\paragraph{\%Inlier.}
We use the inlier rate, \textit{i.e.}, the ratio of inliers to all matches, to evaluate the correspondence quality.
Here, matches whose distance to the ground-truth epipolar line is smaller than certain threshold in both images are regarded as inliers.
To avoid the comparability issue caused by different image resolutions, we set the threshold as $\alpha \sqrt{h^2 + w^2}$, where $h$ and $w$ are height and width of images, respectively.
$\alpha$ is $0.003$ in our evaluation.
Besides, for analyzing intermediate results, we also report \textbf{\%Inlier-m}, \textit{i.e.}, the inlier rate before outlier rejection by robust estimators such as  RANSAC~\cite{fischler1981random}.
This reflects the performance of a pure feature matching system.

\paragraph{\#Corrs.}
We use correspondence numbers for analyzing results rather than performance comparison,
since the impact of match numbers to high-level applications such as SfM~\cite{schonberger2016structure} are arguable~\cite{schonberger2017comparative}.
However, too few correspondences would degenerate these applications.
Therefore, we pay little attention to match numbers, as long as they are not too small.
Similarly, \textbf{\#Corrs-m}, match numbers before the estimation phase, is also reported.

\section{Datasets}~\label{sec-datasets}

We use four large-scale benchmark datasets for evaluation, where different real-world scenes are captured, and camera configurations vary from one to another.
Such diversities allow us to compare algorithms in different scenarios.

\begin{figure}[t]
\centering
\caption{Sample images from the benchmark datasets.}
\includegraphics[width=0.8\textwidth]{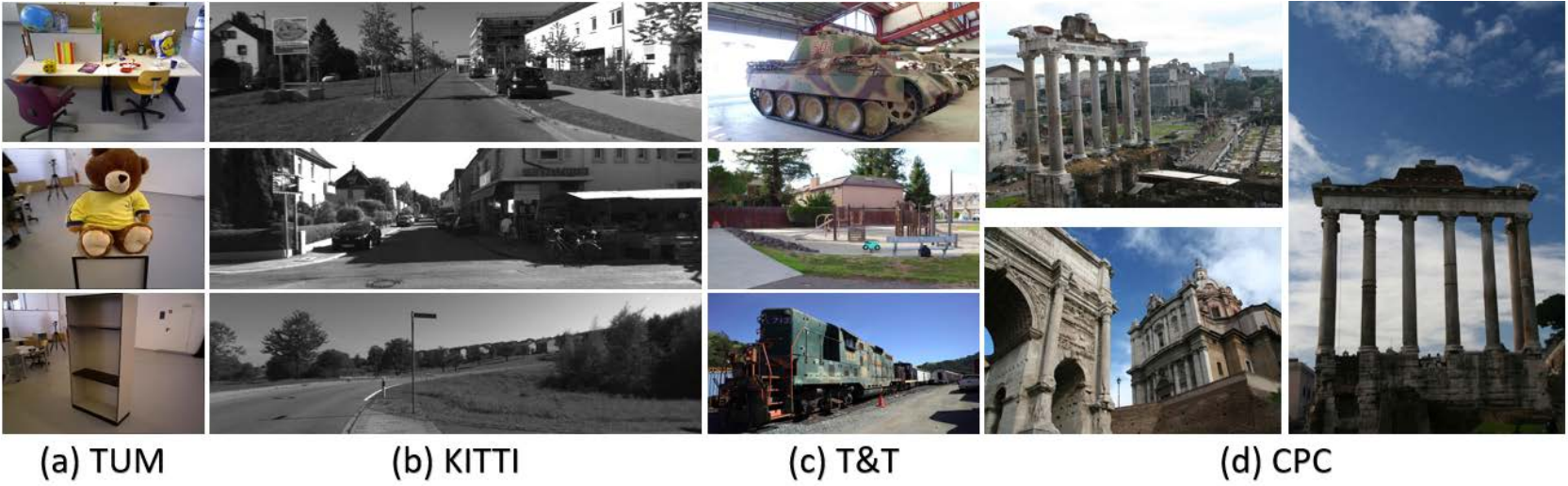}
\label{fig-dataset}
\end{figure}

\paragraph{Datasets.}
The benchmark datasets include:
(1) The TUM SLAM dataset~\cite{sturm12iros}, which provides videos of indoor scenes,
where the texture is often weak and images are sometimes blurred due to the fast camera movement.
(2) The KITTI odometry dataset~\cite{Geiger2012CVPR}, which consists of consecutive frames in a driving scenario, where the geometry between images is dominated by the forward motion.
(3) The Tanks and Temples (T\&T) dataset~\cite{knapitsch2017tanks}, which provides many scans of scenes or objects for image-based reconstruction, and hence offers wide-baseline pairs for evaluation.
(4) The Community Photo Collection (CPC) dataset~\cite{wilson2014robust}, which provides unstructured images of well-known landmarks across the world collected from Flickr.
In the CPC dataset, images are taken from arbitrary cameras at a different time.
\figref{fig-dataset} provides sample images of these benchmark datasets.

\paragraph{Ground Truth.}
The \emph{fundamental matrix} between an image pair could be derived algebraically from their \emph{projection matrices} ($\textbf{P}$ and $\textbf{P}'$) as follows:
\begin{equation}
\textbf{F} = [\textbf{P}'\textbf{C}]_{\times} \textbf{P}'\textbf{P}^+
\end{equation}
where $\textbf{P}^+$ is the pseudo-inverse of $\textbf{P}$, \ie, $\textbf{PP}^+ = \textbf{I}$, and $\textbf{C}$ is a null vector, namely the camera center, defined by $\textbf{PC} = \textbf{0}$.
$\textbf{P} = \textbf{K} [\textbf{R} | \textbf{t}] $ is a 3x4 matrix, and it satisfies 
\begin{equation}
d  \begin{pmatrix} u \\ v \\ 1 \end{pmatrix}  = \textbf{P} \begin{pmatrix} x \\ y \\ z \\ 1 \end{pmatrix}
\end{equation}
where $d$ is an unknown depth, [$u$, $v$] is the image coordinates, and [$x$, $y$, $z$] is the real-world coordinates.
$\textbf{K}$ is the camera intrinsics, and $[\textbf{R} | \textbf{t}]$ is the camera extrinsics.
The ground-truth camera intrinsic and extrinsic parameters are provided in TUM and KITTI datasets, while they are unknown in T\&T and CPC datasets.
Therefore, we derive ground-truth camera parameters for them by reconstructing image sequences using the COLMAP~\cite{schonberger2016structure}, as in~\cite{ranftl2018deep,yi2018learning}.
Note that SfM pipeline reasons globally about the consistency of 3D points and cameras, leading to accurate estimates with an average reprojection error below one pixel~\cite{schonberger2016structure}. 

\paragraph{Image Pairs Construction.}
We search for matchable image pairs by identifying inlier numbers, \ie, we generate correspondences across two images using SIFT~\cite{lowe2004distinctive} and choose pairs which contain more than $20$ inliers, as in~\cite{ranftl2018deep}.
For wide-baseline datasets (T\&T~\cite{knapitsch2017tanks} and CPC~\cite{wilson2014robust}), all image pairs are searched. For short-baseline datasets (TUM~\cite{sturm12iros} and KITTI~\cite{Geiger2012CVPR}), a frame is paired to the subsequent frames captured within one second because almost other pairs are of no overlap.
In this way, we obtain a large number of matchable image pairs, and we randomly choose $1000$ pairs in each dataset for testing.
The testing split on each dataset is described as follows.
In the TUM~\cite{sturm12iros} dataset, we test methods on three sequences: \emph{fr3/teddy}, \emph{fr3/large\_cabinet}, and \emph{fr3/long\_office\_household}.
In the KITTI~\cite{Geiger2012CVPR} Odometry dataset, sequences 06-10 are used.
In the T\&T~\cite{knapitsch2017tanks} dataset, sequences \emph{Panther}, \emph{Playground}, and \emph{Train} are used.
In the CPC~\cite{wilson2014robust} dataset, \emph{Roman Forum} is used.
Other image sequences could be used as training data for further deep learning based methods.
\tabref{tab-dataset} summarizes the test set that we use for evaluation.

\begin{table}[t]
\centering
\caption{Details of the benchmark datasets.}
\resizebox{0.82\textwidth}{!}{
\begin{tabular}{  l | c | c | c | c | r }
   \hline
   Datasets & \#Seq & \#Image & Resolution & Baseline & Property \\
   \hline
   TUM    &3  & 5994  & $480 \times 640$ & short &indoor scenes\\
   \hline
   KITTI  &5  & 9065 & $370\times1226$ & short & street views \\
   \hline
   \multirow{ 2}{*}{T\&T} & \multirow{ 2}{*}{3} & \multirow{ 2}{*}{922} & $1080\times 2048$ & \multirow{ 2}{*}{wide} & \multirow{ 2}{*}{outdoor scenes} \\
   & & & $1080\times 1920$ & &\\
   \hline
   CPC &1 &1615 & varying & wide &internet photos\\
   \hline
\end{tabular}%
}
\label{tab-dataset}
\end{table}

\section{Experiments}~\label{sec-experiments}

Related research is quite rich, so we mainly focus on evaluating recently proposed algorithms and the widely used methods in this paper.
In the following, we introduce the experimental configuration, discuss results, and propose our methods. 

\subsection{Experimental Setup}

\paragraph{Baseline and Comparability.}
We set a classic pipeline as the baseline. 
Specifically, we use DoG~\cite{lowe2004distinctive} detector and SIFT~\cite{lowe2004distinctive} descriptor to generate initial correspondences across images by the plain nearest-neighbor search, then prune bad correspondences using Lowe's \emph{ratio test}~\cite{lowe2004distinctive}, and finally compute FM estimates and remove outliers using RANSAC~\cite{fischler1981random} with the 8-point algorithm~\cite{hartley1997defense}.
For each evaluated algorithm, we incorporate it into the baseline system by replacing its counterpart, and use the overall performance for comparison.

\paragraph{Evaluated Methods.}
Firstly, we evaluate four deep learning based local features, including HesAffNet~\cite{via2018repeatability} detector and two descriptors (L2Net~\cite{tian2017l2}, HardNet++~\cite{mishchuk2017working}).
Besides, two hand-crafted descriptors (DSP-SIFT~\cite{dong2015domain} and RootSIFT-PCA~\cite{arandjelovic2012three,bursuc2015kernel}) are also evaluated, which show high performance in the recent benchmark~\cite{schonberger2017comparative}.
Secondly, we evaluate four correspondence pruning algorithms: CODE~\cite{lin2017code}, GMS~\cite{bian2017gms}, LPM~\cite{Ma2018}, and LC~\cite{yi2018learning}.
Finally, we evaluate two widely used estimators (LMedS~\cite{rousseeuw2005robust} and MSAC~\cite{torr2000mlesac}) and two state-of-the-art alternatives (USAC~\cite{raguram2013usac} and GC-RANSAC~\cite{barath2018graph}).

\paragraph{Implementations.}
We use VLFeat~\cite{vedaldi2010vlfeat} library for the implementation of SIFT descriptor and DoG detector,
and the threshold is $0.8$ for \emph{ratio test}~\cite{lowe2004distinctive}.
Matlab functions are used for RANSAC, LMedS, and MSAC implementations, where we limit the maximum iteration as $2000$ for a reasonable speed.
Other codes are from authors' publicly available implementation, 
where we use the pre-trained models released by authors for deep learning based methods.

\renewcommand{\arraystretch}{1.5}
\begin{table}
\centering
\caption{Experimental results. \textcolor{red}{First}, \textcolor{green}{second}, \textcolor{blue}{third} best results are highlighted in color, and the results that are better than the baseline (the first line in each block) performance are highlighted in bold. \%Recall represents the overall performance.}
\resizebox{1.0\textwidth}{!}{%
\begin{tabular}{c|l c c c r | l c c c r | l c c c r}
\hline
\multirow{ 2}{*}{Datasets} & \multicolumn{5}{c|}{(a) Local Features} & \multicolumn{5}{c|}{(b) Pruning Methods} & \multicolumn{5}{c}{(c) Robust Estimators}\\ 
\cline{2-16}
& Methods &\%Recall &\%Inlier &\%Inlier-m &\#Corrs (-m) & Methods &\%Recall &\%Inlier &\%Inlier-m &\#Corrs (-m) & Methods &\%Recall &\%Inlier &\%Inlier-m &\#Corrs (-m) \\
\hline
\multirow{ 7}{*}{TUM}& SIFT & 57.40& 75.33& 59.21& 65 (316)& RATIO & 57.40& 75.33& 59.21& 65 (316)& RANSAC & \textcolor{green}{57.40}& \textcolor{red}{75.33}& \textcolor{red}{59.21}& 65 (316)\\
& DSP-SIFT & 53.90& 74.89& 56.44& 66 (380)& GMS & \textcolor{green}{\textbf{59.20}}& \textcolor{green}{\textbf{76.18}}& \textcolor{green}{\textbf{69.72}}& 64 (241)& LMedS & \textcolor{red}{\textbf{69.20}}& \textcolor{green}{75.24}& \textcolor{green}{59.21}& 158 (316)\\
& RootSIFT-PCA & \textcolor{red}{\textbf{58.90}}& \textcolor{blue}{\textbf{75.65}}& \textcolor{red}{\textbf{62.22}}& 67 (306)& LPM & \textcolor{blue}{\textbf{58.90}}& \textbf{75.75}& \textbf{64.42}& 67 (290)& MSAC & 52.70& \textcolor{blue}{75.12}& \textcolor{blue}{59.21}& 63 (316)\\
& L2Net & \textcolor{blue}{\textbf{58.10}}& \textbf{75.49}& \textbf{59.26}& 66 (319)& LC & 54.10& \textcolor{blue}{\textbf{75.96}}& \textcolor{red}{\textbf{71.32}}& 57 (203)& USAC & \textcolor{blue}{56.50}& 72.13& 59.21& 244 (316)\\
& HardNet++ & \textcolor{green}{\textbf{58.90}}& \textcolor{red}{\textbf{75.74}}& \textcolor{green}{\textbf{62.07}}& 67 (315) & CODE & \textcolor{red}{\textbf{62.50}}& \textcolor{red}{\textbf{76.95}}& \textcolor{blue}{\textbf{66.82}}& 3119 (18562)& GC-RSC & 30.80& 68.13& 59.21& 272 (316)\\
& HesAffNet & 51.70& \textcolor{green}{\textbf{75.70}}& \textcolor{blue}{\textbf{62.06}}& 101 (657)& & & & & & & & & & \\
\hline
\multirow{ 7}{*}{KITTI}& SIFT & 91.70& 98.20& 87.40& 154 (525)& RATIO & \textcolor{green}{91.70}& 98.20& 87.40& 154 (525)& RANSAC & \textcolor{blue}{91.70}& \textcolor{green}{98.20}& \textcolor{red}{87.40}& 154 (525)\\
& DSP-SIFT & \textcolor{red}{\textbf{92.00}}& \textcolor{green}{\textbf{98.22}}& \textbf{87.60}& 153 (572)& GMS & \textcolor{blue}{91.70}& \textcolor{green}{\textbf{98.58}}& \textcolor{green}{\textbf{95.56}}& 148 (445)& LMedS & \textcolor{red}{\textbf{91.80}}& \textcolor{red}{\textbf{98.25}}& \textcolor{green}{87.40}& 263 (525)\\
& RootSIFT-PCA & \textcolor{green}{\textbf{92.00}}& \textcolor{red}{\textbf{98.23}}& \textcolor{green}{\textbf{90.76}}& 156 (514)& LPM & 91.50& \textbf{98.27}& \textbf{92.50}& 157 (501)& MSAC & \textcolor{green}{\textbf{91.80}}& \textcolor{blue}{98.12}& \textcolor{blue}{87.40}& 153 (525)\\
& L2Net & 91.60& \textcolor{blue}{\textbf{98.21}}& \textbf{89.40}& 156 (520)& LC & 89.70& \textcolor{red}{\textbf{99.44}}& \textcolor{red}{\textbf{97.49}}& 96 (267)& USAC & 82.70& 97.39& 87.40& 455 (525)\\
& HardNet++ & \textcolor{blue}{\textbf{92.00}}& \textbf{98.21}& \textcolor{red}{\textbf{91.25}}& 159 (535) & CODE & \textcolor{red}{\textbf{92.50}}& \textcolor{blue}{\textbf{98.32}}& \textcolor{blue}{\textbf{93.03}}& 4834 (19246)& GC-RSC & 56.50& 95.00& 87.40& 487 (525)\\
& HesAffNet & 90.40& 98.09& \textcolor{blue}{\textbf{90.64}}& 233 (1182)& & & & & & & & & & \\
\hline
\multirow{ 7}{*}{T\&T}& SIFT & 70.00& 75.20& 53.25& 85 (795)& RATIO & 70.00& 75.20& 53.25& 85 (795)& RANSAC & 70.00& 75.20& \textcolor{green}{53.25}& 85 (795)\\
& DSP-SIFT & \textbf{75.10}& \textbf{80.20}& \textbf{60.02}& 90 (845)& GMS & \textcolor{green}{\textbf{80.90}}& \textcolor{green}{\textbf{84.38}}& \textcolor{red}{\textbf{77.65}}& 90 (598)& LMedS & \textcolor{red}{\textbf{83.40}}& \textcolor{blue}{\textbf{77.26}}& \textcolor{blue}{53.25}& 398 (795)\\
& RootSIFT-PCA & \textcolor{blue}{\textbf{77.40}}& \textcolor{blue}{\textbf{80.55}}& \textcolor{blue}{\textbf{61.75}}& 89 (738)& LPM & \textcolor{blue}{\textbf{80.70}}& \textbf{81.62}& \textbf{66.98}& 90 (667)& MSAC & 64.60& 73.27& \textcolor{red}{\textbf{53.43}}& 84 (799)\\
& L2Net & \textbf{70.40}& 73.76& \textbf{57.31}& 93 (799)& LC & \textbf{76.60}& \textcolor{blue}{\textbf{84.01}}& \textcolor{blue}{\textbf{72.24}}& 77 (512)& USAC & \textcolor{blue}{\textbf{78.80}}& \textcolor{red}{\textbf{80.98}}& 53.25& 495 (795)\\
& HardNet++ & \textcolor{green}{\textbf{79.90}}& \textcolor{green}{\textbf{81.05}}& \textcolor{green}{\textbf{63.61}}& 96 (814) & CODE & \textcolor{red}{\textbf{89.40}}& \textcolor{red}{\textbf{89.14}}& \textcolor{green}{\textbf{76.98}}& 782 (9251)& GC-RSC & \textcolor{green}{\textbf{80.40}}& \textcolor{green}{\textbf{78.97}}& 53.25& 612 (795)\\
& HesAffNet & \textcolor{red}{\textbf{82.50}}& \textcolor{red}{\textbf{84.71}}& \textcolor{red}{\textbf{70.29}}& 97 (920)& & & & & & & & & & \\
\hline
\multirow{ 7}{*}{CPC}& SIFT & 29.20& 67.14& 48.07& 60 (415)& RATIO & 29.20& 67.14& 48.07& 60 (415)& RANSAC & 29.20& 67.14& \textcolor{red}{48.07}& 60 (415)\\
& DSP-SIFT & \textbf{35.20}& \textbf{76.48}& \textbf{56.29}& 57 (367)& GMS & \textcolor{green}{\textbf{43.00}}& \textcolor{green}{\textbf{85.90}}& \textcolor{red}{\textbf{82.37}}& 59 (249)& LMedS & \textcolor{blue}{\textbf{44.00}}& \textcolor{blue}{\textbf{75.38}}& \textcolor{green}{48.07}& 209 (415)\\
& RootSIFT-PCA & \textcolor{blue}{\textbf{38.20}}& \textcolor{green}{\textbf{78.45}}& \textcolor{blue}{\textbf{59.92}}& 62 (361)& LPM & \textcolor{blue}{\textbf{39.40}}& \textbf{78.17}& \textbf{65.98}& 60 (310)& MSAC & 23.00& 62.28& \textcolor{blue}{48.07}& 59 (415)\\
& L2Net & \textbf{29.60}& 60.22& \textbf{50.70}& 93 (433)& LC & \textbf{39.40}& \textcolor{blue}{\textbf{83.99}}& \textcolor{blue}{\textbf{72.22}}& 51 (295)& USAC & \textcolor{green}{\textbf{49.70}}& \textcolor{green}{\textbf{80.38}}& 48.07& 232 (415)\\
& HardNet++ & \textcolor{green}{\textbf{40.30}}& \textcolor{blue}{\textbf{76.73}}& \textcolor{green}{\textbf{62.30}}& 69 (400) & CODE & \textcolor{red}{\textbf{51.00}}& \textcolor{red}{\textbf{90.16}}& \textcolor{green}{\textbf{78.55}}& 696 (5774)& GC-RSC & \textcolor{red}{\textbf{53.70}}& \textcolor{red}{\textbf{81.15}}& 48.07& 269 (415)\\
& HesAffNet & \textcolor{red}{\textbf{47.40}}& \textcolor{red}{\textbf{84.58}}& \textcolor{red}{\textbf{72.22}}& 65 (405)& & & & & & & & & & \\
\hline
\end{tabular}% 
}
\label{tab-results}
\end{table}

\subsection{Results and Discussion}

\tabref{tab-results} reports the experimental results for all methods and datasets.
In each block, the first line shows the baseline performance.
\textcolor{red}{First}, \textcolor{green}{second}, \textcolor{blue}{third} best results are highlighted in color,
and the results that are better than the baseline are highlighted in bold.
Here, we mainly compare algorithms in terms of \emph{\%Recall}, which reflects the overall performance.
In addition, \emph{\%Inlier} shows matching performance, and \emph{\%Inlier-m} shows matching before outlier rejection phase.
\emph{\#Corrs(-m)} is used to analyze results instead of performance comparison.
The detail about these metrics can be seen in~\secref{sec-metrics}. 
For performance analyses, we mainly target on concluding the distinctive properties of the best methods instead of a comprehensive comparison of all approaches.

\paragraph{Local Features.}
\tabref{tab-results}(a) shows the results of local features.
The \emph{\%Recall} implies that
\textbf{a)} RootSIFT-PCA~\cite{bursuc2015kernel} and HardNet++~\cite{mishchuk2017working} consistently outperform the baseline, and the latter is better than the former.
\textbf{b)} HesAffNet~\cite{via2018repeatability} performs best in wide-baseline scenarios (T\&T and CPC), although it is degenerate on the TUM dataset.
\textbf{c)} DSP-SIFT~\cite{dong2015domain} outperforms the baseline on almost all datasets but TUM, and L2Net~\cite{tian2017l2} shows similar performances with the baseline on all datasets.

\paragraph{Correspondence Pruning Methods.}
\tabref{tab-results}(b) shows the results of pruning methods.
It shows that
\textbf{a)} CODE~\cite{lin2017code} achieves the state-of-the-art performance on all datasets.
\textbf{b)} GMS~\cite{bian2017gms} and LPM~\cite{Ma2018} consistently outperform the baseline methods by pruning bad correspondences effectively, \ie, improving the \emph{\%Inlier-m} and preserving a considerable \emph{\#Corrs-m} in the meanwhile.
Here, GMS is better than LPM.
\textbf{c)} LC~\cite{yi2018learning} can boost the matching accuracy (\%Inlier-m) but, for estimation (\emph{\%Recall}), it is degenerate on the short-baseline datasets (TUM and KITTI).
Perhaps this is because the provided model is trained on wide-baseline datasets.
Also, note that it requires camera intrinsics, which are normally assumed to be unknown for the FM estimation problem.

\paragraph{Robust Estimators.}
\tabref{tab-results}(c) shows the results of robust estimators.
They show:
\textbf{a)} LMedS~\cite{rousseeuw2005robust} performs best on the first three datasets where images are not as difficult as CPC dataset.
This confirms the suggestion by Matlab documentation that LMedS works well when the inlier rate is high enough, \eg, above $50\%$.
\textbf{b)} GC-RANSAC~\cite{barath2018graph} and USAC~\cite{raguram2013usac} show high performances in wide-baseline scenarios, especially on the challenging CPC dataset.
However, they are degenerate in short-baseline scenarios (TUM and KITTI).
\textbf{c)} Interestingly, we observe that GC-RANSAC (also USAC) can preserve rich correspondences (\emph{\%Corrs-m}) and prune outliers (\emph{\%Inlier(-m)}) effectively.

\paragraph{Runtime.}
As algorithms rely on different operating systems, we use two machines for evaluation: a Linux server \textbf{L} (Intel E5-2620 CPU, NVIDIA Titan Xp GPU) and a Windows laptop \textbf{W} ( Intel i7-3630QM CPU, NVIDIA GeForce GT 650M GPU), where $100$ images from the KITTI dataset are used for testing and the averaged results are reported.
\tabref{table-timecost} reports the time consumption of algorithms.
Descriptors rely on DoG~\cite{lowe2004distinctive} detector, which (\textbf{L}) takes $238ms$ to extract $1760$ keypoints, and HesAffNet~\cite{via2018repeatability} detector extracts $4860$ keypoints.
CODE~\cite{lin2017code} (\textbf{W}) takes $2.953s$ to extract $58,675$ keypoints using GPU, and takes $4.068s$ to prune bad correspondences using CPU.

% Third, note that SIFT variants can be accelerated by using GPU in applications.
% Overall, besides CODE, other algorithms are (or could be) efficient in current devices.
% , and RANSAC variants could also be optimized in applications.

\renewcommand{\arraystretch}{1.2}
\begin{table}[t]
\centering
\caption{Time consumption of evaluated algorithms.}
\resizebox{0.8\textwidth}{!}{%
\begin{tabular}{c| c c c c c c c}
\hline
Device & \multicolumn{7}{|c}{Runtime (seconds)} \\
\hline
\multirow{ 2}{*}{L} & SIFT & DSP-SIFT & RootSIFT-PCA & L2Net & HardNet++ & HesAffNet\\
& 0.702 & 1.762 & 0.705 & 2.260  & 0.002 & 0.367\\
\hline
\multirow{ 4}{*}{W} & LPM & GMS & LC & CODE & & &\\
& 0.003 & 0.001 & 0.021 & 4.068 & & &\\
\cline{2-8}
& RANSAC & LMedS &MSAC &USAC &GC-RSC & & \\
& 0.521 & 0.528 & 0.537 &0.565 & 0.788 & &\\ 
\hline
\end{tabular}%
}
\label{table-timecost}
\end{table}

\subsection{Proposed Methods}

Drawing inspiration from the results, we propose three practical matching systems and a robust estimator as follows.

\paragraph{Matching Systems.}
We first adopt one of the following three pairs of detectors and descriptors for generating putative correspondences:
\begin{enumerate}
	\setlength\itemsep{0em}
	\item DoG~\cite{lowe2004distinctive} + RootSIFT-PCA~\cite{bursuc2015kernel}
    \item DoG + (HardNet++)~\cite{mishchuk2017working}
    \item HesAffNet~\cite{via2018repeatability} + (HardNet++)
\end{enumerate}
where we recommend 1, 2 for general scenes and 3 for wide-baseline scenarios.
Then, we apply \emph{ratio test} (the threshold is $0.8$) and GMS~\cite{bian2017gms} to prune bad correspondences.
Finally, we use LMedS~\cite{rousseeuw2005robust} based estimator for model fitting.
\tabref{tab-recommend} shows the evaluation results, which clearly demonstrate that the recommended systems outperform the baseline, and achieve competitive performances with the state-of-the-art system (CODE~\cite{lin2017code} + LMedS~\cite{rousseeuw2005robust}).
Note that CODE is several orders of magnitude slower, even GPU is adopted.

\paragraph{Coarse-to-Fine RANSAC.}

\tabref{tab-results} shows that GC-RANSAC~\cite{barath2018graph} and USAC~\cite{raguram2013usac} prune outliers effectively, although they fail to show consistently high performance on model fitting.
To this end, we propose to use GC-RANSAC~\cite{barath2018graph} for pruning bad matches, and then apply LMedS~\cite{rousseeuw2005robust} based estimator for model fitting.
Note that USAC is also applicable.
In this two-stage framework, the former is used to roughly find the inlier set and the latter to fit the model accurately, so we term the resultant approach Coarse-to-Fine RANSAC (CF-RSC in short).
\tabref{tab-cfransac} shows the results of the proposed method in terms of \emph{\%Recall}, where all estimators use the same input, \ie, SIFT~\cite{lowe2004distinctive} matches with \emph{ratio test} pruning.
It shows that the proposed CF-RSC significantly outperforms other alternatives.
% The proposed CF-RSC can be used in a wide range of computer vision tasks for robustness enhancement, especially where the input data is noisy.
% For data points of a high inlier rate, \eg, the correspondences outputted by the recommended matching systems, CF-RSC achieves very similar performances with LMedS~\cite{rousseeuw2005robust} estimator.

\begin{table}[t]
\centering
\caption{Evaluation results of the proposed matching systems.}
\resizebox{0.8\textwidth}{!}{%
\begin{tabular}{l| l |c| c| c| r}
\hline
Datasets & Methods &\%Recall &\%Inlier &\%Inlier-m &\#Corrs(-m)\\
\hline
\multirow{ 5}{*}{TUM} & Baseline  & 57.40& 75.33& 59.21& 65 (316)\\
&CODE  & \textcolor{blue}{\textbf{67.50}}& \textcolor{green}{\textbf{76.04}}& \textbf{66.82}& 9281 (18562)\\
&RootSIFT-PCA + GMS  & \textcolor{green}{\textbf{67.50}}& \textcolor{red}{\textbf{76.13}}& \textcolor{red}{\textbf{69.62}}& 124 (248)\\
&HardNet + GMS  & \textcolor{red}{\textbf{68.60}}& \textbf{75.85}& \textcolor{green}{\textbf{69.39}}& 128 (256)\\
&HesAffNet + GMS  & \textbf{66.40}& \textcolor{blue}{\textbf{75.92}}& \textcolor{blue}{\textbf{67.04}}& 288 (577)\\
\hline
\multirow{ 5}{*}{KITTI} & Baseline  & 91.70& 98.20& 87.40& 154 (525)\\
&CODE  & \textcolor{blue}{\textbf{91.90}}& \textbf{98.22}& \textbf{93.03}& 9623 (19246)\\
&RootSIFT-PCA + GMS  & \textcolor{red}{\textbf{92.50}}& \textcolor{red}{\textbf{98.54}}& \textcolor{red}{\textbf{95.73}}& 225 (450)\\
&HardNet + GMS  & \textcolor{green}{\textbf{92.10}}& \textcolor{green}{\textbf{98.49}}& \textcolor{green}{\textbf{95.43}}& 236 (472)\\
&HesAffNet + GMS  & \textbf{91.80}& \textcolor{blue}{\textbf{98.48}}& \textcolor{blue}{\textbf{94.18}}& 540 (1079)\\
\hline
\multirow{ 5}{*}{T\&T} & Baseline  & 70.00& 75.20& 53.25& 85 (795)\\
&CODE  & \textcolor{red}{\textbf{92.70}}& \textcolor{red}{\textbf{87.81}}& \textbf{76.98}& 4626 (9251)\\
&RootSIFT-PCA + GMS  & \textbf{89.30}& \textbf{85.29}& \textcolor{blue}{\textbf{78.69}}& 307 (614)\\
&HardNet + GMS  & \textcolor{green}{\textbf{92.20}}& \textcolor{blue}{\textbf{85.52}}& \textcolor{green}{\textbf{78.86}}& 343 (686)\\
&HesAffNet + GMS  & \textcolor{blue}{\textbf{90.90}}& \textcolor{green}{\textbf{86.16}}& \textcolor{red}{\textbf{79.25}}& 412 (824)\\
\hline
\multirow{ 5}{*}{CPC} & Baseline  & 29.20& 67.14& 48.07& 60 (415)\\
&CODE  & \textcolor{red}{\textbf{61.80}}& \textcolor{red}{\textbf{89.45}}& \textbf{78.55}& 2890 (5774)\\
&RootSIFT-PCA + GMS  & \textbf{57.30}& \textcolor{green}{\textbf{88.94}}& \textcolor{red}{\textbf{83.70}}& 133 (263)\\
&HardNet + GMS  & \textcolor{blue}{\textbf{60.10}}& \textbf{88.34}& \textcolor{blue}{\textbf{83.12}}& 149 (298)\\
&HesAffNet + GMS  & \textcolor{green}{\textbf{60.80}}& \textcolor{blue}{\textbf{88.72}}& \textcolor{green}{\textbf{83.16}}& 182 (362)\\
\hline
\end{tabular}%
}
\label{tab-recommend}
\end{table}

\begin{table}[t]
\centering
\caption{\%Recall of the proposed CF-RSC.}
\resizebox{0.8\textwidth}{!}{%
\begin{tabular}{ l | c | c | c | c | c | c }
\hline
Datasets &RANSAC &LMedS &MSAC &USAC &GC-RSC &CF-RSC \\
\hline
TUM & 57.40 & 69.20 & 52.70 & 56.50 & 30.80 & \textbf{69.30} \\ \hline
KITTI & 91.70 & 91.80 & 91.80 & 82.70 & 56.50 & \textbf{92.30} \\ \hline
T\&T & 70.00 & 83.40 & 64.60 & 78.80 & 80.40 & \textbf{90.70} \\ \hline
CPC & 29.20 & 44.00 & 23.00 & 49.70 & 53.70 & \textbf{60.90} \\ \hline
\end{tabular}%\
}
\label{tab-cfransac}
\end{table}

\section{Conclusions}~\label{sec-conclusions}

This paper evaluates the recently proposed local features, correspondence pruning algorithms, and robust estimators using strictly defined metrics in the context of image matching and fundamental matrix estimation.
Comprehensive evaluation results on four large-scale datasets provide insights into which datasets are particularly challenging and which algorithms perform well in which scenarios.
This can advance the development of related research fields, 
and it can also help researchers design practical matching systems in different applications.
Finally, drawing inspiration from the results, we propose three high-quality image matching systems and a robust estimator, Coarse-to-Fine RANSAC.
They achieve remarkable performances and have potentials in a wide range of computer vision tasks.

\section{Acknowledgement}

The authors would like to thank TuSimple and Huawei Technologies Co. Ltd.

\bibliography{reference}

\begin{thebibliography}{48}
\providecommand{\natexlab}[1]{#1}
\providecommand{\url}[1]{\texttt{#1}}
\expandafter\ifx\csname urlstyle\endcsname\relax
  \providecommand{\doi}[1]{doi: #1}\else
  \providecommand{\doi}{doi: \begingroup \urlstyle{rm}\Url}\fi

\bibitem[Agarwal et~al.(2011)Agarwal, Furukawa, Snavely, Simon, Curless, Seitz,
  and Szeliski]{agarwal2011building}
Sameer Agarwal, Yasutaka Furukawa, Noah Snavely, Ian Simon, Brian Curless,
  Steven~M Seitz, and Richard Szeliski.
\newblock Building {Rome} in a day.
\newblock \emph{Communications of the ACM}, 54\penalty0 (10):\penalty0
  105--112, 2011.

\bibitem[Arandjelovic and Zisserman(2012)]{arandjelovic2012three}
Relja Arandjelovic and Andrew Zisserman.
\newblock Three things everyone should know to improve object retrieval.
\newblock In \emph{{IEEE Conference on Computer Vision and Pattern Recognition
  (CVPR)}}, pages 2911--2918. IEEE, 2012.

\bibitem[Armangu{\'e} and Salvi(2003)]{armangue2003overall}
Xavier Armangu{\'e} and Joaquim Salvi.
\newblock Overall view regarding fundamental matrix estimation.
\newblock \emph{Image and Vision Computing}, 21\penalty0 (2):\penalty0
  205--220, 2003.

\bibitem[Balntas et~al.(2016)Balntas, Riba, Ponsa, and
  Mikolajczyk]{balntas2016learning}
Vassileios Balntas, Edgar Riba, Daniel Ponsa, and Krystian Mikolajczyk.
\newblock Learning local feature descriptors with triplets and shallow
  convolutional neural networks.
\newblock In \emph{{British Machine Vision Conference (BMVC)}}, page~3, 2016.

\bibitem[Balntas et~al.(2017)Balntas, Lenc, Vedaldi, and
  Mikolajczyk]{hpatches_2017_cvpr}
Vassileios Balntas, Karel Lenc, Andrea Vedaldi, and Krystian Mikolajczyk.
\newblock {HPatches}: A benchmark and evaluation of handcrafted and learned
  local descriptors.
\newblock In \emph{{IEEE Conference on Computer Vision and Pattern Recognition
  (CVPR)}}, pages 5173--5182. IEEE, 2017.

\bibitem[Barath and Matas(2018)]{barath2018graph}
Daniel Barath and Jiri Matas.
\newblock {Graph-Cut} {RANSAC}.
\newblock In \emph{{IEEE Conference on Computer Vision and Pattern Recognition
  (CVPR)}}, 2018.

\bibitem[Bian et~al.(2017)Bian, Lin, Matsushita, Yeung, Nguyen, and
  Cheng]{bian2017gms}
JiaWang Bian, Wen-Yan Lin, Yasuyuki Matsushita, Sai-Kit Yeung, Tan~Dat Nguyen,
  and Ming-Ming Cheng.
\newblock {GMS}: Grid-based motion statistics for fast, ultra-robust feature
  correspondence.
\newblock In \emph{{IEEE Conference on Computer Vision and Pattern Recognition
  (CVPR)}}, pages 4181--4190. IEEE, 2017.

\bibitem[Brown et~al.(2011)Brown, Hua, and Winder]{brown2011discriminative}
Matthew Brown, Gang Hua, and Simon Winder.
\newblock Discriminative learning of local image descriptors.
\newblock \emph{{IEEE Transactions on Pattern Recognition and Machine
  Intelligence (PAMI)}}, pages 43--57, 2011.

\bibitem[Bursuc et~al.(2015)Bursuc, Tolias, and J{\'e}gou]{bursuc2015kernel}
Andrei Bursuc, Giorgos Tolias, and Herv{\'e} J{\'e}gou.
\newblock Kernel local descriptors with implicit rotation matching.
\newblock In \emph{International Conference on Multimedia Retrieval}, pages
  595--598. ACM, 2015.

\bibitem[Choi et~al.(1997)Choi, Kim, and Yu]{choi1997performance}
Sunglok Choi, Taemin Kim, and Wonpil Yu.
\newblock Performance evaluation of {RANSAC} family.
\newblock \emph{{International Journal on Computer Vision (IJCV)}}, 24\penalty0
  (3):\penalty0 271--300, 1997.

\bibitem[Davison et~al.(2007)Davison, Reid, Molton, and
  Stasse]{davison2007monoslam}
Andrew~J Davison, Ian~D Reid, Nicholas~D Molton, and Olivier Stasse.
\newblock {MonoSLAM}: Real-time single camera slam.
\newblock \emph{{IEEE Transactions on Pattern Recognition and Machine
  Intelligence (PAMI)}}, 29\penalty0 (6):\penalty0 1052--1067, 2007.

\bibitem[Dong and Soatto(2015)]{dong2015domain}
Jingming Dong and Stefano Soatto.
\newblock Domain-size pooling in local descriptors: {DSP-SIFT}.
\newblock In \emph{{IEEE Conference on Computer Vision and Pattern Recognition
  (CVPR)}}, pages 5097--5106, 2015.

\bibitem[Fathy et~al.(2011)Fathy, Hussein, and Tolba]{fathy2011fundamental}
Mohammed~E Fathy, Ashraf~S Hussein, and Mohammed~F Tolba.
\newblock Fundamental matrix estimation: A study of error criteria.
\newblock \emph{Pattern Recognition Letters}, 32\penalty0 (2):\penalty0
  383--391, 2011.

\bibitem[Fischler and Bolles(1981)]{fischler1981random}
Martin~A Fischler and Robert~C Bolles.
\newblock Random sample consensus: a paradigm for model fitting with
  applications to image analysis and automated cartography.
\newblock \emph{Communications of the ACM}, 24\penalty0 (6):\penalty0 381--395,
  1981.

\bibitem[Forster et~al.(2014)Forster, Pizzoli, and Scaramuzza]{forster2014svo}
Christian Forster, Matia Pizzoli, and Davide Scaramuzza.
\newblock {SVO}: Fast semi-direct monocular visual odometry.
\newblock In \emph{{International Conference on Robotics and Automation
  (ICRA)}}, pages 15--22. IEEE, 2014.

\bibitem[Geiger et~al.(2012)Geiger, Lenz, and Urtasun]{Geiger2012CVPR}
Andreas Geiger, Philip Lenz, and Raquel Urtasun.
\newblock Are we ready for autonomous driving? the {KITTI} vision benchmark
  suite.
\newblock In \emph{{IEEE Conference on Computer Vision and Pattern Recognition
  (CVPR)}}, pages 3354--3361. IEEE, 2012.

\bibitem[Hartley and Zisserman(2003)]{hartley2003multiple}
Richard Hartley and Andrew Zisserman.
\newblock \emph{Multiple view geometry in computer vision}.
\newblock Cambridge university press, 2003.

\bibitem[Hartley(1997)]{hartley1997defense}
Richard~I Hartley.
\newblock In defense of the eight-point algorithm.
\newblock \emph{{IEEE Transactions on Pattern Recognition and Machine
  Intelligence (PAMI)}}, 19\penalty0 (6):\penalty0 580--593, 1997.

\bibitem[Heinly et~al.(2012)Heinly, Dunn, and Frahm]{heinly2012comparative}
Jared Heinly, Enrique Dunn, and Jan-Michael Frahm.
\newblock Comparative evaluation of binary features.
\newblock In \emph{{European Conference on Computer Vision (ECCV)}}, pages
  759--773. Springer, 2012.

\bibitem[Heinly et~al.(2015)Heinly, Schonberger, Dunn, and
  Frahm]{heinly2015reconstructing}
Jared Heinly, Johannes~L Schonberger, Enrique Dunn, and Jan-Michael Frahm.
\newblock Reconstructing the world* in six days*(as captured by the yahoo 100
  million image dataset).
\newblock In \emph{{IEEE Conference on Computer Vision and Pattern Recognition
  (CVPR)}}, pages 3287--3295, 2015.

\bibitem[Knapitsch et~al.(2017)Knapitsch, Park, Zhou, and
  Koltun]{knapitsch2017tanks}
Arno Knapitsch, Jaesik Park, Qian-Yi Zhou, and Vladlen Koltun.
\newblock Tanks and {Temples}: Benchmarking large-scale scene reconstruction.
\newblock \emph{{ACM Transactions on Graphics (TOG)}}, 36\penalty0
  (4):\penalty0 78, 2017.

\bibitem[Lacey et~al.(2000)Lacey, Pinitkarn, and Thacker]{lacey2000evaluation}
AJ~Lacey, N~Pinitkarn, and Neil~A Thacker.
\newblock An evaluation of the performance of {RANSAC} algorithms for stereo
  camera calibrarion.
\newblock In \emph{{British Machine Vision Conference (BMVC)}}, pages 1--10,
  2000.

\bibitem[Lin et~al.(2017)Lin, Wang, Cheng, Yeung, Torr, Do, and
  Lu]{lin2017code}
Wen-Yan Lin, Fan Wang, Ming-Ming Cheng, Sai-Kit Yeung, Philip~HS Torr, Minh~N
  Do, and Jiangbo Lu.
\newblock {CODE}: Coherence based decision boundaries for feature
  correspondence.
\newblock \emph{{IEEE Transactions on Pattern Recognition and Machine
  Intelligence (PAMI)}}, 2017.

\bibitem[Lowe(2004)]{lowe2004distinctive}
David~G Lowe.
\newblock Distinctive image features from scale-invariant keypoints.
\newblock \emph{{International Journal on Computer Vision (IJCV)}}, 60\penalty0
  (2):\penalty0 91--110, 2004.

\bibitem[Luo et~al.(2018)Luo, Shen, Zhou, Zhu, Zhang, Yao, Fang, and
  Quan]{luo2018geodesc}
Zixin Luo, Tianwei Shen, Lei Zhou, Siyu Zhu, Runze Zhang, Yao Yao, Tian Fang,
  and Long Quan.
\newblock {GeoDesc}: Learning local descriptors by integrating geometry
  constraints.
\newblock In \emph{{European Conference on Computer Vision (ECCV)}}, 2018.

\bibitem[Ma et~al.(2018)Ma, Zhao, Jiang, Zhou, and Guo]{Ma2018}
Jiayi Ma, Ji~Zhao, Junjun Jiang, Huabing Zhou, and Xiaojie Guo.
\newblock Locality preserving matching.
\newblock \emph{{International Journal on Computer Vision (IJCV)}}, 2018.

\bibitem[Mikolajczyk and Schmid(2005)]{mikolajczyk2005performance}
Krystian Mikolajczyk and Cordelia Schmid.
\newblock A performance evaluation of local descriptors.
\newblock \emph{{IEEE Transactions on Pattern Recognition and Machine
  Intelligence (PAMI)}}, 27\penalty0 (10):\penalty0 1615--1630, 2005.

\bibitem[Mikolajczyk et~al.(2005)Mikolajczyk, Tuytelaars, Schmid, Zisserman,
  Matas, Schaffalitzky, Kadir, and Van~Gool]{mikolajczyk2005comparison}
Krystian Mikolajczyk, Tinne Tuytelaars, Cordelia Schmid, Andrew Zisserman, Jiri
  Matas, Frederik Schaffalitzky, Timor Kadir, and Luc Van~Gool.
\newblock A comparison of affine region detectors.
\newblock \emph{{International Journal on Computer Vision (IJCV)}}, 65\penalty0
  (1-2):\penalty0 43--72, 2005.

\bibitem[Mishchuk et~al.(2017)Mishchuk, Mishkin, Radenovic, and
  Matas]{mishchuk2017working}
Anastasiia Mishchuk, Dmytro Mishkin, Filip Radenovic, and Jiri Matas.
\newblock Working hard to know your neighbor's margins: Local descriptor
  learning loss.
\newblock In \emph{{Neural Information Processing Systems (NIPS)}}, pages
  4826--4837, 2017.

\bibitem[Mishkin et~al.(2018)Mishkin, Radenovic, and
  Matas]{via2018repeatability}
Dmytro Mishkin, Filip Radenovic, and Jiri Matas.
\newblock Repeatability is not enough: Learning affine regions via
  discriminability.
\newblock In \emph{{European Conference on Computer Vision (ECCV)}}. Springer,
  2018.

\bibitem[Mur-Artal et~al.(2015)Mur-Artal, Montiel, and Tardos]{mur2015orb}
Raul Mur-Artal, Jose Maria~Martinez Montiel, and Juan~D Tardos.
\newblock {ORB-SLAM}: a versatile and accurate monocular slam system.
\newblock \emph{{IEEE Transactions on Robotics (TRO)}}, 31\penalty0
  (5):\penalty0 1147--1163, 2015.

\bibitem[Radenovic et~al.(2016)Radenovic, Sch{\"o}nberger, Ji, Frahm, Chum, and
  Matas]{radenovic2016dusk}
Filip Radenovic, Johannes~L Sch{\"o}nberger, Dinghuang Ji, Jan-Michael Frahm,
  Ondrej Chum, and Jiri Matas.
\newblock From dusk till dawn: Modeling in the dark.
\newblock In \emph{{IEEE Conference on Computer Vision and Pattern Recognition
  (CVPR)}}, pages 5488--5496, 2016.

\bibitem[Raguram et~al.(2008)Raguram, Frahm, and
  Pollefeys]{raguram2008comparative}
Rahul Raguram, Jan-Michael Frahm, and Marc Pollefeys.
\newblock A comparative analysis of {RANSAC} techniques leading to adaptive
  real-time random sample consensus.
\newblock In \emph{{European Conference on Computer Vision (ECCV)}}, pages
  500--513. Springer, 2008.

\bibitem[Raguram et~al.(2013)Raguram, Chum, Pollefeys, Matas, and
  Frahm]{raguram2013usac}
Rahul Raguram, Ondrej Chum, Marc Pollefeys, Jiri Matas, and Jan-Michael Frahm.
\newblock {USAC}: a universal framework for random sample consensus.
\newblock \emph{{IEEE Transactions on Pattern Recognition and Machine
  Intelligence (PAMI)}}, 35\penalty0 (8):\penalty0 2022--2038, 2013.

\bibitem[Ranftl and Koltun(2018)]{ranftl2018deep}
Ren{\'e} Ranftl and Vladlen Koltun.
\newblock Deep fundamental matrix estimation.
\newblock In \emph{{European Conference on Computer Vision (ECCV)}}, pages
  284--299, 2018.

\bibitem[Rousseeuw and Leroy(1987)]{rousseeuw2005robust}
Peter~J Rousseeuw and Annick~M Leroy.
\newblock \emph{Robust regression and outlier detection}, volume 589.
\newblock John wiley \& sons, 1987.

\bibitem[Sch{\"o}nberger and Frahm(2016)]{schonberger2016structure}
Johannes~L Sch{\"o}nberger and Jan-Michael Frahm.
\newblock Structure-from-motion revisited.
\newblock In \emph{{IEEE Conference on Computer Vision and Pattern Recognition
  (CVPR)}}, pages 4104--4113, 2016.

\bibitem[Sch{\"o}nberger et~al.(2015)Sch{\"o}nberger, Radenovic, Chum, and
  Frahm]{schonberger2015single}
Johannes~L Sch{\"o}nberger, Filip Radenovic, Ondrej Chum, and Jan-Michael
  Frahm.
\newblock From single image query to detailed 3d reconstruction.
\newblock In \emph{{IEEE Conference on Computer Vision and Pattern Recognition
  (CVPR)}}, pages 5126--5134, 2015.

\bibitem[Sch{\"o}nberger et~al.(2017)Sch{\"o}nberger, Hardmeier, Sattler, and
  Pollefeys]{schonberger2017comparative}
Johannes~L Sch{\"o}nberger, Hans Hardmeier, Torsten Sattler, and Marc
  Pollefeys.
\newblock Comparative evaluation of hand-crafted and learned local features.
\newblock In \emph{{IEEE Conference on Computer Vision and Pattern Recognition
  (CVPR)}}, pages 6959--6968. IEEE, 2017.

\bibitem[Sturm et~al.(2012)Sturm, Engelhard, Endres, Burgard, and
  Cremers]{sturm12iros}
J.~Sturm, N.~Engelhard, F.~Endres, W.~Burgard, and D.~Cremers.
\newblock A benchmark for the evaluation of {RGB-D SLAM} systems.
\newblock In \emph{{IEEE International Conference on Intelligent Robots and
  Systems (IROS)}}, Oct. 2012.

\bibitem[Tian et~al.(2017)Tian, Fan, Wu, et~al.]{tian2017l2}
Yurun Tian, Bin Fan, Fuchao Wu, et~al.
\newblock {L2-Net}: Deep learning of discriminative patch descriptor in
  euclidean space.
\newblock In \emph{{IEEE Conference on Computer Vision and Pattern Recognition
  (CVPR)}}, volume~1, page~6, 2017.

\bibitem[Torr and Murray(1997)]{torr1997development}
Philip~HS Torr and David~W Murray.
\newblock The development and comparison of robust methods for estimating the
  fundamental matrix.
\newblock \emph{{International Journal on Computer Vision (IJCV)}}, 24\penalty0
  (3):\penalty0 271--300, 1997.

\bibitem[Torr and Zisserman(2000)]{torr2000mlesac}
Philip~HS Torr and Andrew Zisserman.
\newblock {MLESAC}: A new robust estimator with application to estimating image
  geometry.
\newblock \emph{{Computer Vision and Image Understanding (CVIU)}}, 78\penalty0
  (1):\penalty0 138--156, 2000.

\bibitem[Torr and Zissermann(1997)]{torr1997performance}
Philip~HS Torr and A~Zissermann.
\newblock Performance characterization of fundamental matrix estimation under
  image degradation.
\newblock \emph{Machine Vision and Applications}, 9\penalty0 (5-6):\penalty0
  321--333, 1997.

\bibitem[Vedaldi and Fulkerson(2010)]{vedaldi2010vlfeat}
Andrea Vedaldi and Brian Fulkerson.
\newblock {VLFeat}: An open and portable library of computer vision algorithms.
\newblock In \emph{{ACM International Conference on Multimedia (ACM MM)}},
  pages 1469--1472. ACM, 2010.

\bibitem[Wilson and Snavely(2014)]{wilson2014robust}
Kyle Wilson and Noah Snavely.
\newblock Robust global translations with {1DSFM}.
\newblock In \emph{{European Conference on Computer Vision (ECCV)}}, pages
  61--75. Springer, 2014.

\bibitem[Yi et~al.(2018)Yi, Trulls, Ono, Lepetit, Salzmann, and
  Fua]{yi2018learning}
Kwang~Moo Yi, Eduard Trulls, Yuki Ono, Vincent Lepetit, Mathieu Salzmann, and
  Pascal Fua.
\newblock Learning to find good correspondences.
\newblock In \emph{{IEEE Conference on Computer Vision and Pattern Recognition
  (CVPR)}}, 2018.

\bibitem[Zhang(1998)]{Zhang1998}
Zhengyou Zhang.
\newblock Determining the epipolar geometry and its uncertainty: A review.
\newblock \emph{{International Journal on Computer Vision (IJCV)}}, 27\penalty0
  (2):\penalty0 161--195, Mar 1998.
\newblock ISSN 1573-1405.
\newblock \doi{10.1023/A:1007941100561}.

\end{thebibliography}
\end{document}

\maketitle

\section{Matching pipeline overview}

\noindent
\textbf{Feature Extraction.}
In image matching problem, the first step is extracting local features, which involves finding interest points from images using \emph{detectors}~\cite{harris1988combined,lowe2004distinctive,via2018repeatability}, and representing their local features using \emph{descriptors}~\cite{lowe2004distinctive,luo2018geodesc,mishchuk2017working,tian2017l2}.
The quality of detectors and descriptors is crucial to the overall system, as which are starting points and main primitives for subsequent algorithms.

\vspace{2mm}
\noindent
\textbf{Descriptor Matching.}
Exhaustively finding the nearest neighbors for the detected features allows initial correspondences across two images.
As it is slow in real applications, many approximated search techniques are proposed~\cite{muja2009fast,cheng2014fast}.
They aim to reduce the search complexity while preserving the matching accuracy in the meanwhile.
Besides, the \emph{vocabulary tree}~\cite{fei2005bayesian} is widely used to match one image to a large database, known as the image retrieval problem~\cite{philbin2007object}.
In this paper, we use the plain nearest-neighbor method implemented in GPU for accuracy and efficiency.

\vspace{2mm}
\noindent
\textbf{Correspondence Pruning.}
False correspondences degenerate matching performance significantly.
Perhaps the most widely used algorithm to prune them is Lowe's \emph{ratio test}~\cite{lowe2004distinctive}, which discards descriptors whose top-ranked two nearest neighbors are very similar.
Besides, the constraints in photometry~\cite{liu2012virtual}, locality~\cite{Ma2018}, motion consistency~\cite{bian2017gms}, and global coherence~\cite{lin2017code} are studied to distinguish true matches from false matches.
Recently, a deep learning based approach~\cite{yi2018learning} is proposed for this task.
It works like a model fitting method, \ie, finding good correspondences by verifying whether they satisfy the estimated \emph{essential matrix}.

\vspace{2mm}
\noindent
\textbf{Robust FM Estimation.}
The fundamental matrix (FM) could be recovered from outlier-free data by using the 8-point algorithm~\cite{hartley1997defense} or the 7-point algorithm~\cite{hartley2003multiple}.
Then various measures can be taken to deal with outliers.
The most popular method, to our knowledge, is RANSAC~\cite{fischler1981random}, which searches for a model that has the most support using random sampling.
A large amount of literatures on variations of this basic idea exist, \eg, the classic LMedS~\cite{rousseeuw2005robust} and MSAC~\cite{torr2000mlesac}, the state-of-the-art USAC~\cite{raguram2013usac}, and the very recent GC-RANSAC~\cite{barath2018graph}.
Besides, deep learning techniques are studied to tackle this problem recently~\cite{ranftl2018deep}.

\section{Runtime}

\vspace{2mm}
\noindent
\textbf{Runtime}.
Since evaluated algorithms rely on different operating systems, we use two machines: a Linux server \textbf{L} (Intel E5-2620 CPU, NVIDIA Titan Xp GPU) and a Windows laptop \textbf{W} ( Intel i7-3630QM CPU, NVIDIA GeForce GT 650M GPU), where $100$ images from the KITTI dataset are used for testing and the averaged results are reported. 
\tabref{table-timecost} reports the time consumption of algorithms.
First, descriptors rely on DoG~\cite{lowe2004distinctive} detector, which (in \textbf{L}) takes $238ms$ to extract $1760$ keypoints, and HesAffNet~\cite{via2018repeatability} detector extracts $4860$ keypoints.
Second, CODE~\cite{lin2017code} (in \textbf{W}) takes $2.953s$ to extract $58,675$ keypoints using GPU, and takes $4.068s$ to prune bad correspondences in CPU.
Third, note that SIFT variants can be accelerated by GPU, and RANSAC variants could also be optimized in applications.
Overall, besides CODE, other algorithms are (or could be) efficient in current devices.

\renewcommand{\arraystretch}{1.2}
\begin{table}[ht]
\centering
\resizebox{0.8\textwidth}{!}{%
\begin{tabular}{|c| c c c c c c c|}
\hline
Device & \multicolumn{7}{|c|}{Runtime (seconds)} \\
\hline
\multirow{ 2}{*}{L} & SIFT & DSP-SIFT & RootSIFT-PCA & L2Net & GeoDesc & HardNet++ & HesAffNet\\
& 0.702 & 1.762 & 0.705 & 2.260 & 0.094 & 0.002 & 0.367\\
\hline
\multirow{ 4}{*}{W} & LPM & GMS & LC & CODE & & &\\
& 0.003 & 0.001 & 0.021 & 4.068 & & &\\
\cline{2-8}
& RANSAC & LMedS &MSAC &USAC &GC-RSC & & \\
& 0.521 & 0.528 & 0.537 &0.565 & 0.788 & &\\ 
\hline
\end{tabular}%
}
\vspace{0.2mm}
\caption{Time consumption of evaluated algorithms.}
\label{table-timecost}
\vspace{-3mm}
\end{table}

% \begin{figure}[t]
% \centering
% \resizebox{0.6\textwidth}{!}{%
% \begin{tabular}{cc}
% 	\includegraphics[width=1\linewidth]{recall.pdf} &
%     \includegraphics[width=1\linewidth]{inliers.pdf} \\
% \end{tabular}%
% }
% \vspace{0.2mm}
% \caption{Baseline performance on the benchmark datasets. \%Recall shows the estimation performance, and we report the results on $0.05$ accuracy threshold for a numerical comparison. \%Inlier(-m) shows the matching performance after(-before) outlier removal by RANSAC.
% }
% \label{fig-baseline}
% \vspace{-3mm}
% \end{figure}

\section{Details about evaluated methods}
\vspace{2mm}
\noindent
\textbf{Evaluated Local Features}.
We evaluate four deep learning based features, including one detector (HesAffNet~\cite{via2018repeatability}) and three descriptors (L2Net~\cite{tian2017l2}, HardNet++~\cite{mishchuk2017working}, and GeoDesc~\cite{luo2018geodesc}).
Besides, two hand-crafted descriptors (DSP-SIFT~\cite{dong2015domain} and RootSIFT-PCA~\cite{arandjelovic2012three,bursuc2015kernel}) are also evaluated.
Following~\cite{schonberger2017comparative}, we use DoG~\cite{lowe2004distinctive} detector to extract interest points for all descriptors and, for HesAffNet~\cite{via2018repeatability} detector, we use HardNet++~\cite{mishchuk2017working} descriptor as authors recommended.
Here, we use the SIFT (DoG) implementation by VLFeat~\cite{vedaldi2010vlfeat}, and use open-source codes and provided models by authors for deep methods, where the best results are reported when multiple models are given.
For matching descriptors, we use the plain nearest-neighbor matching method implemented in GPU and prune initial correspondences using \emph{ratio test}~\cite{lowe2004distinctive}, where the threshold is $0.8$, as in Heinly benchmark~\cite{heinly2012comparative}.
Note that although authors suggest $0.89$ for GeoDesc~\cite{luo2018geodesc}, we empirically find it degenerates the performance in our evaluation.

\vspace{2mm}
\noindent
\textbf{Evaluated Pruning Methods}.
We evaluate four correspondence pruning algorithms: CODE~\cite{lin2017code}, GMS~\cite{bian2017gms}, LPM~\cite{Ma2018}, and LC~\cite{yi2018learning}.
In these algorithms, GMS and LPM do not require additional information.
CODE depends on ASIFT feature~\cite{morel2009asift}, and LC requires camera intrinsics, which are unknown in real-world FM estimation problem.
We use COLMAP~\cite{schonberger2016structure} computed camera intrinsics for LC.
For GMS, LPM, and LC, we use SIFT~\cite{lowe2004distinctive} with \emph{ratio test} to generate correspondences, where the ratio threshold and hyperparameters are tunned by maximizing the estimation recall.
The results show that the ratio should be $0.8$ for LPM~\cite{Ma2018} and GMS~\cite{bian2017gms}, and be $0.9$ for LC~\cite{yi2018learning}.
Besides, for GMS, the multi-scale mode is activated for wide-baseline scenarios (on the T\&T and CPC datasets), and the threshold factor is $4$ in our evaluation.
We empirically find that this is better than using the default configuration provided by authors.
For CODE~\cite{lin2017code}, we regard it as a complete matching system, and the publicly available binary program is evaluated.

\vspace{2mm}
\noindent
\textbf{Evaluated Robust Estimators}.
We evaluate two widely used methods: LMedS~\cite{rousseeuw2005robust} and MSAC~\cite{torr2000mlesac};
and two state-of-the-art alternatives: USAC~\cite{raguram2013usac} and GC-RANSAC~\cite{barath2018graph}.
Here, the RANSAC, LMedS, and MSAC implementations are from Matlab, in which the 8-point algorithm~\cite{hartley1997defense} is used.
For USAC and GC-RANSAC, we use the publicly available implementations by authors, where the 7-point algorithm~\cite{hartley2003multiple} is adopted.
Note that USAC requires the ranking of input points, so we sort correspondences using their two nearest-neighbor ratios.
Besides, we set the maximum number of iterations to be $2000$ for all methods to ensure a reasonable speed.
As all methods work on the same input, \ie, SIFT~\cite{lowe2004distinctive} matches pruned by \emph{ratio test}, the comparison is fair.

\section{Details about dataset construction}

\vspace{2mm}
\noindent
\textbf{Overlapped Pairs.}
We search for matchable image pairs by identifying inlier numbers, \ie, we generate correspondences across two images using SIFT~\cite{lowe2004distinctive} matcher, and choose pairs who contain more than $20$ inliers, as in~\cite{ranftl2018deep}.
Here, a match would be regarded as inlier, if its distances to the ground-truth epipolar lines is within $1$ pixels in both images.
For wide-baseline datasets (T\&T~\cite{knapitsch2017tanks} and CPC~\cite{wilson2014robust}), all image pairs are searched and, for short-baseline datasets (TUM~\cite{sturm12iros} and KITTI~\cite{Geiger2012CVPR}), a frame is paired to the following frames captured within one second because almost other pairs are of no overlap.
In this way, we obtain a large number of matchable image pairs.
For testing purposes, we randomly choose $1000$ pairs in each dataset.

\vspace{2mm}
\noindent
\textbf{Testing/Training Splits.}
Although only the testing set is used in this paper, we also provide the training set for encouraging future learning based approaches.
Here, we follow the splits by Ranft \etal~\cite{ranftl2018deep} in the last three datasets, where the TUM~\cite{sturm12iros} dataset is not used.
The consistency in dataset splits would facilitate the comparison for future research.
We describe it as follows:
In the TUM~\cite{sturm12iros} dataset, we test methods on three sequences: \emph{fr3/teddy}, \emph{fr3/large\_cabinet}, and \emph{fr3/long\_office\_household}, and leave other sequences for training.
In the KITTI~\cite{Geiger2012CVPR} dataset, sequences 06-10 are used for testing, and sequences 00-05 for training.
In the T\&T~\cite{knapitsch2017tanks} dataset, sequences \emph{Panther}, \emph{Playground}, and \emph{Train} are used for testing, and \emph{Family}, \emph{Francis}, \emph{Horse}, \emph{Lighthouse}, and \emph{M60} for training.
In the CPC~\cite{wilson2014robust} dataset, \emph{Roman Forum} is used for testing and others for training.

\bibliography{reference}